%% file: 0-main.tex
\documentclass[letterpaper]{article} 
\usepackage{aaai25}  
\usepackage{times}  
\usepackage{helvet}  
\usepackage{courier}  
\usepackage[hyphens]{url}  
\usepackage{graphicx} 
\usepackage{natbib}  
\usepackage{caption} 
\usepackage{algorithm}
\usepackage{algorithmic}
\usepackage{amsfonts}
\usepackage{amsmath}
\usepackage{xcolor}
\usepackage{booktabs}
\usepackage{array}
\usepackage{xspace}
\usepackage{pgfplots}
\usepackage{makecell}
\usepackage{multirow}
\usepackage{subcaption}

\newtheorem{problem}{Problem}

\newcommand{\hide}[1]{}
\newcommand{\model}{SeNER\xspace}
\newcommand{\biswa}{BiSPA\xspace}
\newcommand{\tokenspans}{token-pair spans\xspace}
\newcommand{\tokenspan}{token-pair span\xspace}
\newcommand{\profiling}{Scholar-XL\xspace}
\newcommand{\scirex}{SciREX\xspace}
\newcommand{\oldprofiling}{Profiling-07\xspace}

\usepackage{newfloat}
\usepackage{listings}
\DeclareCaptionStyle{ruled}{labelfont=normalfont,labelsep=colon,strut=off} 
\floatstyle{ruled}
\newfloat{listing}{tb}{lst}{}
\floatname{listing}{Listing}
\title{Small Language Model Makes an Effective Long Text Extractor}
\author{
    Yelin Chen\textsuperscript{\rm 1}\equalcontrib \thanks{Part of the work was done when Yelin interned at Zhipu AI.},
    Fanjin Zhang\textsuperscript{\rm 2}\footnotemark[1]\thanks{Corresponding authors.},
    Jie Tang\textsuperscript{\rm 2}\footnotemark[3]
}

\affiliations{
    \textsuperscript{\rm 1}School of Computer Science and Technology, Xinjiang University, Urumqi 830049, China

    \textsuperscript{\rm 2}Department of Computer Science and Technology, Tsinghua University, Beijing 100084, China

    ylin@stu.xju.edu.cn, 
    \{fanjinz, jietang\}@tsinghua.edu.cn
}

\usepackage{bibentry}

\begin{document}

\maketitle

\begin{abstract}
Named Entity Recognition (NER) is a fundamental problem in natural language processing (NLP).
However, the task of extracting longer entity spans (e.g., awards) from extended texts (e.g., homepages) is barely explored.
Current NER methods predominantly fall into two categories: span-based methods and generation-based methods.
Span-based methods require the enumeration of all possible token-pair spans, followed by classification on each span, resulting in substantial redundant computations and excessive GPU memory usage.
In contrast, generation-based methods involve prompting or fine-tuning large language models (LLMs) to adapt to downstream NER tasks.
However, these methods struggle with the accurate generation of longer spans and often incur significant time costs for effective fine-tuning.
To address these challenges, 
this paper introduces a lightweight span-based NER method called \model, 
which incorporates a bidirectional arrow attention mechanism coupled with LogN-Scaling on the \texttt{[CLS]} token to embed long texts effectively,
and comprises a novel bidirectional sliding-window plus-shaped attention (\biswa) mechanism to
reduce redundant candidate token-pair spans significantly and model interactions between token-pair spans simultaneously.
Extensive experiments demonstrate that our method achieves state-of-the-art extraction accuracy on three long NER datasets
and is capable of extracting entities from long texts in a GPU-memory-friendly manner.
\end{abstract}

\begin{links}
    \link{Code}{https://github.com/THUDM/scholar-profiling/tree/main/sener}
\end{links}

%

\input{Introduction}
\input{RelatedWork}
\input{problem}
\input{Method}
\input{Experiment}
\input{Conclusion}

\newpage

\section{Acknowledgments}

This work is supported by NSFC for Distinguished Young Scholar 62425601, Tsinghua University Initiative Scientific Research Program and the New Cornerstone Science Foundation through the XPLORER PRIZE.
This work is also supported by the Natural Science Foundation of China (NSFC) 62406164,
the
Postdoctoral Fellowship Program of CPSF under Grant Number GZB20240358 and 2024M761680.

\bibliography{aaai25}

\appendix
\newpage
\input{appendix}

\end{document}

%% file: Introduction.tex
\section{Introduction}

Named entity recognition (NER), a fundamental task in information extraction (IE), aims to identify spans indicating specific types of entities.
It serves as the foundation for numerous downstream tasks, including relation extraction~\cite{miwa2016end}, knowledge graph construction~\cite{xu2017cn}, and question answering~\cite{molla2006named}.

\begin{figure}[t]
\centering
\includegraphics[width=0.9\columnwidth]{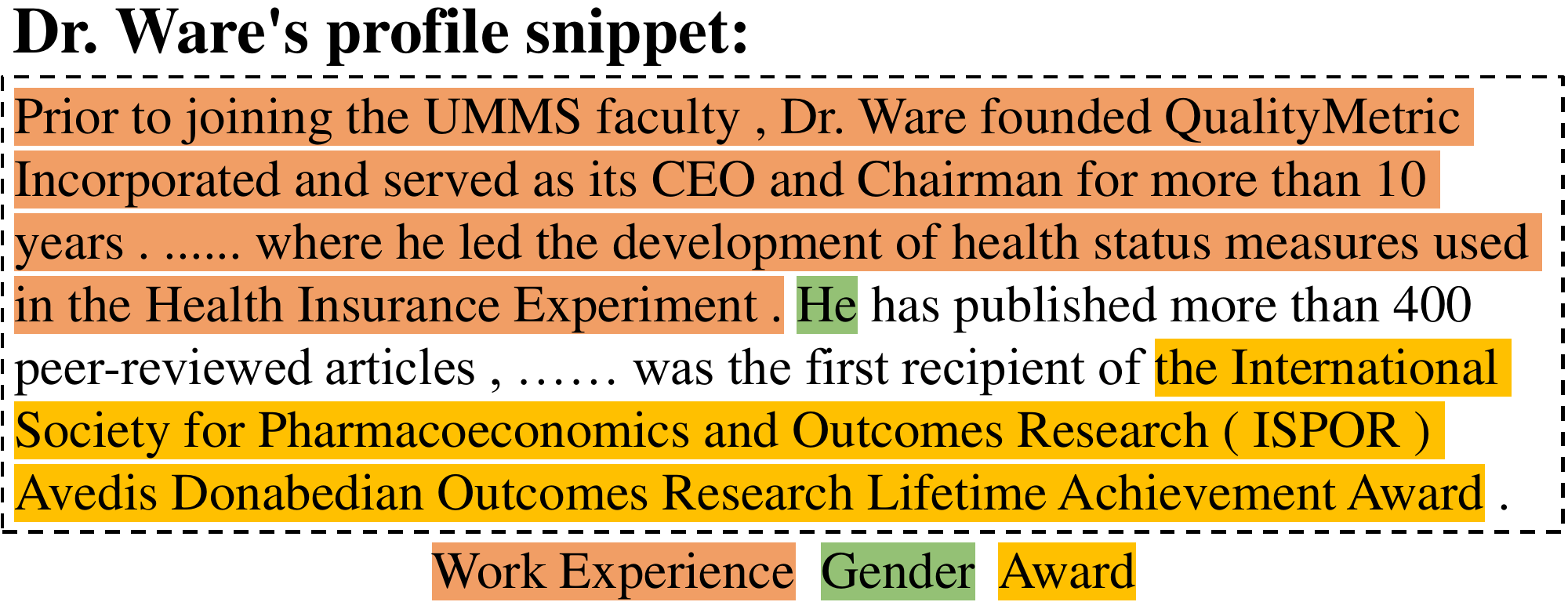} 
\caption{An example of entity/attribute extraction from an author's homepage,
where ``work experience'' is a long entity block
and ``award'' is a long entity.
}
\label{fig:data-example}
\end{figure}

\begin{figure*}[t]
	\begin{subfigure}{0.47\linewidth}
		\centering
		\includegraphics[width=0.9\linewidth]{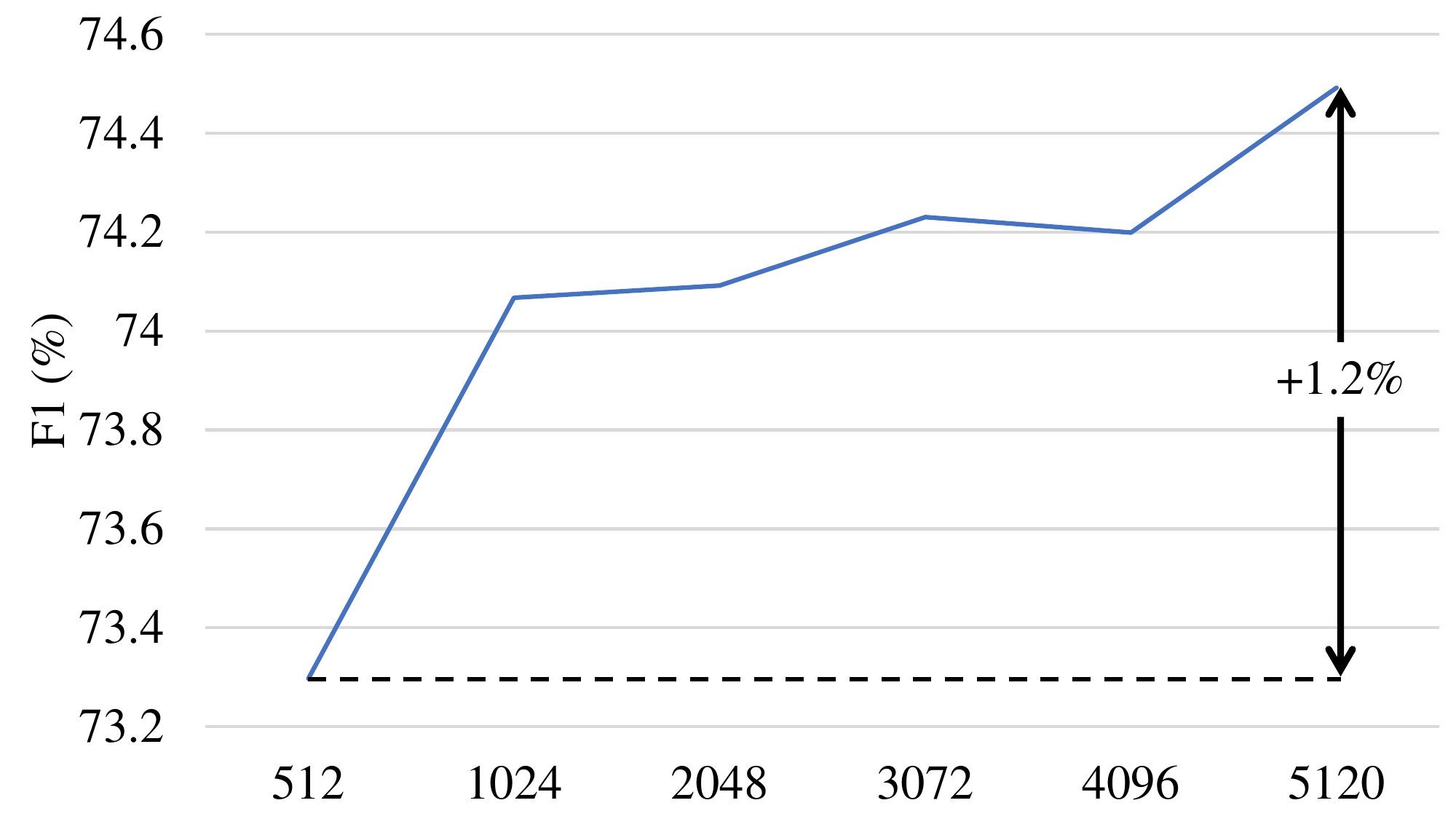}
		\caption{Impact of input length for our model \model regarding extraction performance on the \scirex dataset (\%).}
		\label{fig:length-analysis-scirex}
	\end{subfigure}
    \hspace{0.8cm}
    \begin{subfigure}{0.45\linewidth}
		\includegraphics[width=0.9\linewidth]{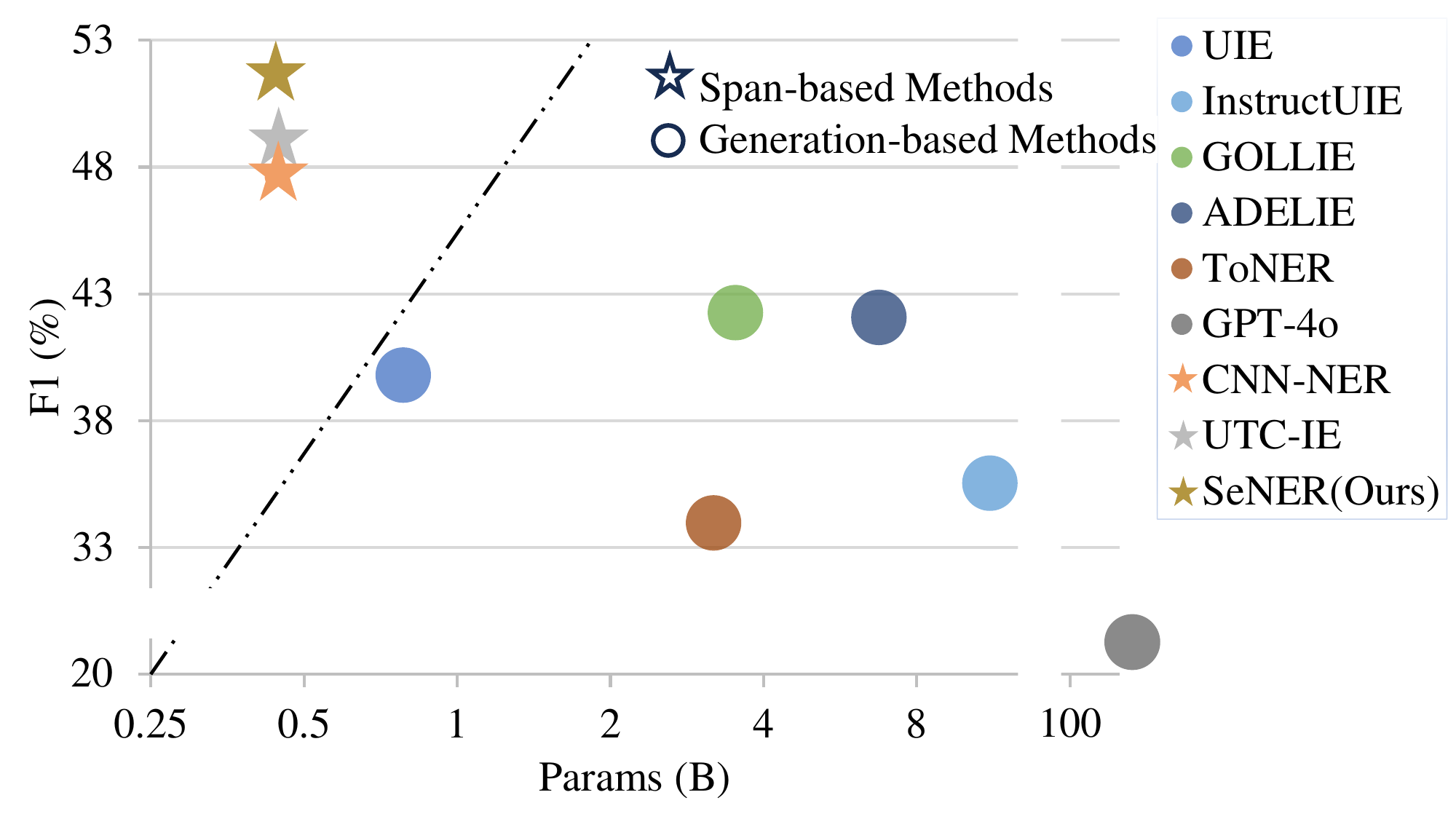}
		\caption{Model parameters of span-based and generation-based methods vs. F1 score on the \profiling dataset.}
		\label{fig:para-vs-f1}
	\end{subfigure}
	\caption{Performance of entity recognition with respect to input length and the number of model parameters of NER methods.}
\end{figure*}

Despite extensive studies, existing NER research rarely focuses on extracting named entities from long texts, a common real-world scenario such as extracting author attributes from homepages and identifying ``methods'' and ``problems'' in academic papers.
For example, in Figure \ref{fig:data-example},
``work experience'' is a long entity block 
while ``award'' is a long entity,
posing greater challenges for the NER task.
We also extend 
the input length of a NER method to extract short entities in academic papers, as shown in 
Figure \ref{fig:length-analysis-scirex},
suggesting that long input length brings clear benefits to extract entities more precisely due to the perception of longer contexts.

Traditional approaches treat the NER task as a sequence labeling task, assigning a single label to each token, exemplified by the BIOES format.
However, these methods are inadequate for recognizing nested entities. 
To address this issue, later efforts typically employ span-based methods~\cite{su2022global,yan-etal-2023-utc}, which consider all possible \tokenspans and classify each span.
These methods achieve satisfactory accuracy on sentence-based NER tasks
but struggle to identify entity blocks across sentences or extract entities from long texts
due to substantial redundant computations and GPU memory usage resulting from $\mathcal{O}(L^2)$ computation complexity based on the \tokenspan tensor,
where $L$ is the input length.

Recently, large language models (LLMs) demonstrate remarkable performance on a spectrum of natural language understanding and generation tasks~\cite{zhao2023survey}.
However, LLMs still fall short and do not align well with information extraction tasks~\cite{qi2024adelie}.
On the Scholar-XL dataset~\cite{zhang2024oag},
extracting author attributes via prompting GPT-4o~\cite{achiam2023gpt} by providing $5$ similar demonstrations only achieves a $21.87\%$ F1 score, as shown in Figure \ref{fig:para-vs-f1}.
Fine-tuning LLMs to extract entities from long texts is also feasible~\cite{sainz2023gollie,qi2024adelie}, but it incurs significant time costs for training and inference compared to span-based methods, without guaranteeing high accuracy.

Therefore, we aim to recognize named entities from long texts in a GPU-memory-friendly way without compromising accuracy.
Given these limitations, 
we propose \model,
a lightweight span-based method to extract entities from long texts.
The main idea of \model is to reduce the redundant computations during the encoding and extraction processes of long texts.
\model presents two core components that have an edge over existing span-based NER methods.

\begin{itemize}
    \item To encode long texts effectively and efficiently,
    we employ a bidirectional arrow attention mechanism that encodes both local and global contexts simultaneously.
    To overcome the entropy instability issue of input texts of varied length, we apply 
    LogN-Scaling~\cite{su2021logn} on the \texttt{[CLS]} token to keep the entropy of attention scores stable.
    \item To reduce superfluous span-based computation and model interactions between \tokenspans, we propose a novel bidirectional sliding-window plus attention (\biswa) mechanism to efficiently compute horizontal and vertical attention on focused spans.
\end{itemize}

To enhance the robustness and generalization of our model,
we employ the whole word masking strategy~\cite{cui2021pre}  and the LoRA~\cite{hu2021lora} technique during training.
Extensive experimental results on three datasets highlight the superiority of our proposed method.
\model achieves state-of-the-art accuracy while maintaining relatively small model parameters, as depicted in Figure \ref{fig:para-vs-f1}.
Additionally, under the same hardware and configuration, our model is capable of handling texts $6$ times longer than previous advanced span-based NER methods.

%% file: RelatedWork.tex
\section{Related Work}

NER methods are generally categorized into span-based methods, generation-based methods, and other methods.

\subsection{Span-based Methods}

Span-based methods~\cite{li2022unified,yuan-etal-2022-fusing,su2022global,zhu2022boundary} reframe the NER task as a \tokenspan
classification task. 
They identify spans based on start and end positions, enumerate all possible candidate spans in a sentence, and perform classification. 
Most existing methods focus on obtaining high-quality span representations and modeling interactions between spans.
CNN-NER~\cite{yan-etal-2023-embarrassingly} utilizes Convolutional Neural Networks (CNNs) to model spatial relations in the \tokenspan tensor. 
UTC-IE~\cite{yan-etal-2023-utc} further incorporates axis-aware interaction with plus-shaped self-attention for the \tokenspan tensor on top of CNN-NER.
These methods offer parallel extraction, simple decoding, and advantages in handling nested entity recognition, leading to widespread use and excellent performance. 
However, calculating all span representations and aggregating interactions between \tokenspans requires substantial computational resources, which limits their effectiveness for long texts. 

\subsection{Generation-based Methods}

Generation-based methods extract entities from text in an end-to-end manner, 
where the generated sequence can be 
text~\cite{lu2022unified,jiang2024toner}, entity pointers~\cite{yan2021unified}, or code~\cite{sainz2023gollie}. 
With the rise of large language models (LLMs), such methods~\cite{wang2023gpt,xie-etal-2023-empirical,ashok2023promptner} achieve good performance with only a few examples due to their generalization abilities. 
Some methods~\cite{wang2023instructuie,dagdelen2024structured} enhance general extraction capabilities by using powerful LLMs, high-quality data, diverse extraction tasks, and comprehensive prior knowledge. 
GoLLIE~\cite{sainz2023gollie} ensures adherence to annotation guidelines through strategies such as class order shuffling, class dropout, guideline paraphrasing, representative candidate sampling, and class name masking. 
ADELIE~\cite{qi2024adelie} performs instruction tuning
on a high-quality alignment corpus and further optimizes it with a Direct Preference Optimization (DPO) objective. 
However, compared to span-based methods,
these methods often require significant computational resources and may perform poorly in generating accurate longish entities from long texts.
The construction of instructions and use of examples can compress input text length, leading to low text utilization. Additionally, autoregressive generation can result in long decoding times. 

\subsection{Other Methods}

In addition to the two main paradigms, there are a few other types of methods. 
Some methods~\cite{ma2016end,yan2019tener,strakova2019neural} model the NER task as a sequence labeling task. 
However, these methods struggle with nested entities. 
Some methods~\cite{li2019unified,tan2021sequence,shen2022parallel} use two independent multi-layer perceptrons (MLPs) to predict the start and end positions of entities separately, which can lead to errors due to treating the entity as separate modules. 
Some approaches~\cite{lou-etal-2022-nested,yang-tu-2022-bottom} employ hypergraphs to represent spans, but their decoding processes is complex.

%% file: problem.tex
\section{Problem Definition}

In this section, we introduce the problem formulation of named entity recognition from long texts.

\begin{problem}
\textbf{Named Entity Recognition from Long Texts (Long NER).}
Given a long input text, 
the goal is to extract different types of named entities or entity blocks
that mark their start and end positions in the text.
\textnormal{Note that the input length can exceed $1{\small ,}000$ tokens and the entity length can exceed $100$ tokens in our problem.}
\end{problem}

Taking scholar profiling~\cite{schiaffino2009intelligent,gu2018profiling} as an example, ``birth place'' is a kind of entity,
while ``work experience'' often appears as an entity block
that involves multiple segments.

%% file: Method.tex
\section{Method}

As previously discussed, conventional NER methods fall into two main categories: span-based and generation-based.
For NER in long texts, span-based methods need to model interactions between \tokenspans, which incurs substantial GPU memory and computation.
In contrast, generation-based methods, 
commonly based on LLMs, are arduous to generate longish entity spans accurately.

\begin{figure}[t]
\centering
\includegraphics[width=0.7\columnwidth]{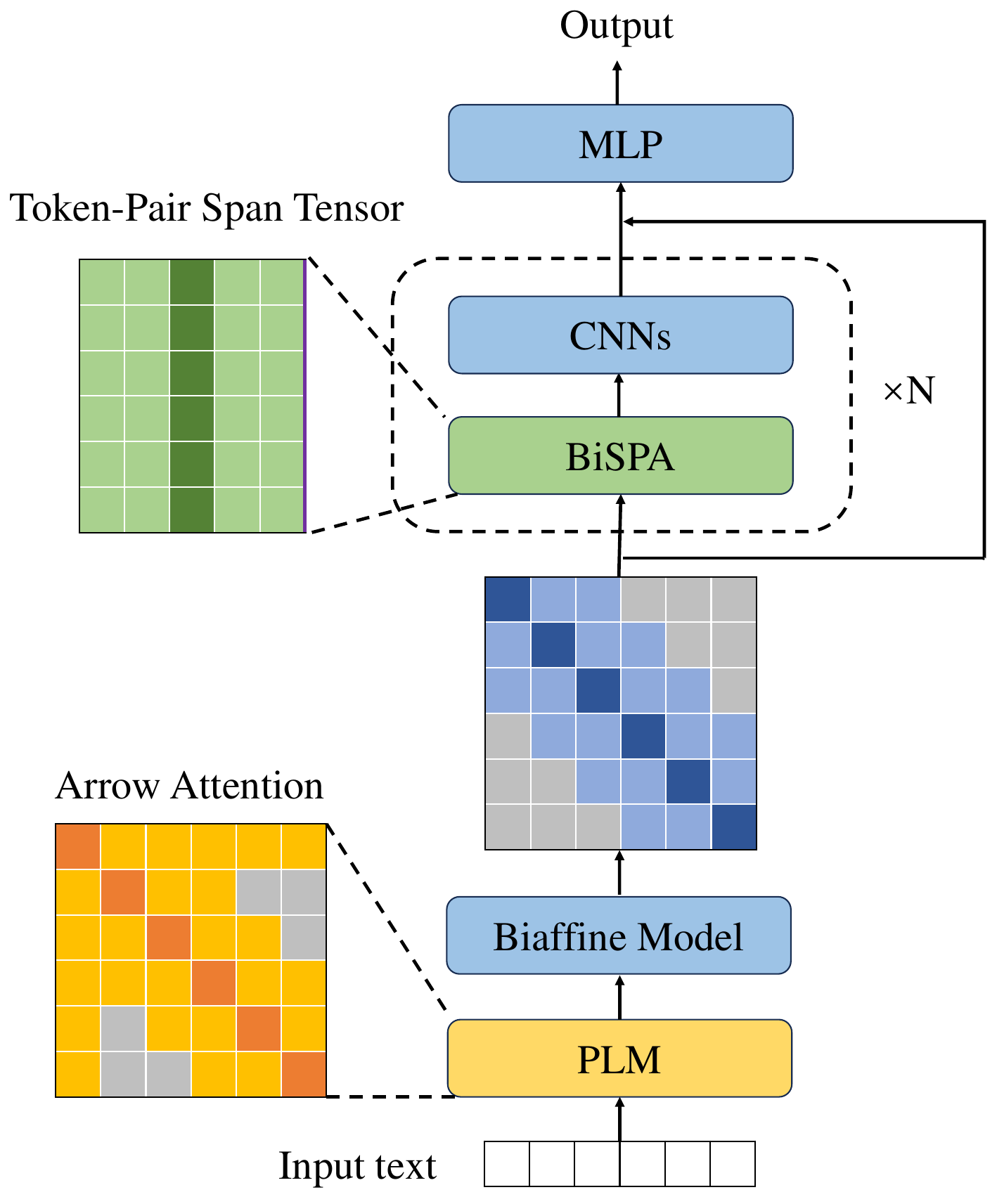} 
\caption{An overview of the \model model.}
\label{fig:model}
\end{figure}

In response to these limitations, we propose a lightweight span-based NER model, \model, that efficiently encodes long input texts and models \tokenspans interactions.
First, we employ a pre-trained language model (PLM) with a arrow attention mechanism to encode long inputs efficiently. 
To alleviate entropy instability resulting from varied input lengths, we apply LogN-Scaling~\cite{su2021logn} to the \texttt{[CLS]} token.
Next, we leverage a Biaffine model~\cite{dozat2017deep} to obtain the hidden representation of each \tokenspan. 
Then, we present the \tokenspan interaction module,
where we propose a novel \biswa mechanism to significantly reduce redundant candidate token pairs
and model interactions between token pairs simultaneously.
Finally, we introduce the training strategy and prediction method.
An overview of our model is shown in Figure \ref{fig:model}.

\subsection{Long Input Encoding}
Given a piece of text, we pass it into a PLM to obtain its contextual vector representation.

\begin{equation}
H=\left [ h_{1}, h_{2}, ..., h_{L} \right ]=\text{PLM}\left ( \left [ x_{1}, x_{2}, ..., x_{L} \right ] \right )\label{eq1}
\end{equation}

\noindent where $H\in \mathbb{R}^{L\times d}$, $L$ is the input length, and $d$ is the output dimension of the PLM. 

Traditional NER methods utilize PLMs with full bidirectional attention, incuring a large amount of GPU memory footprint and computation for long texts.
Moreover, full attention for long texts is often unnecessary since distant tokens are usually semantically unrelated.
In light of this, a straightforward idea is to use sliding window attention (SWA)~\cite{beltagy2020longformer, zaheer2020big},
which adopts a fixed window, say $w$, so that 
each token attends to $w$ tokens to its left and $w$ tokens to its right.
However, SWA ignores the global context, impairing the ability of the Transformer layers to acquire a comprehensive understanding of the entire input text.

\begin{figure}[t]
\centering
\includegraphics[width=0.9\columnwidth]{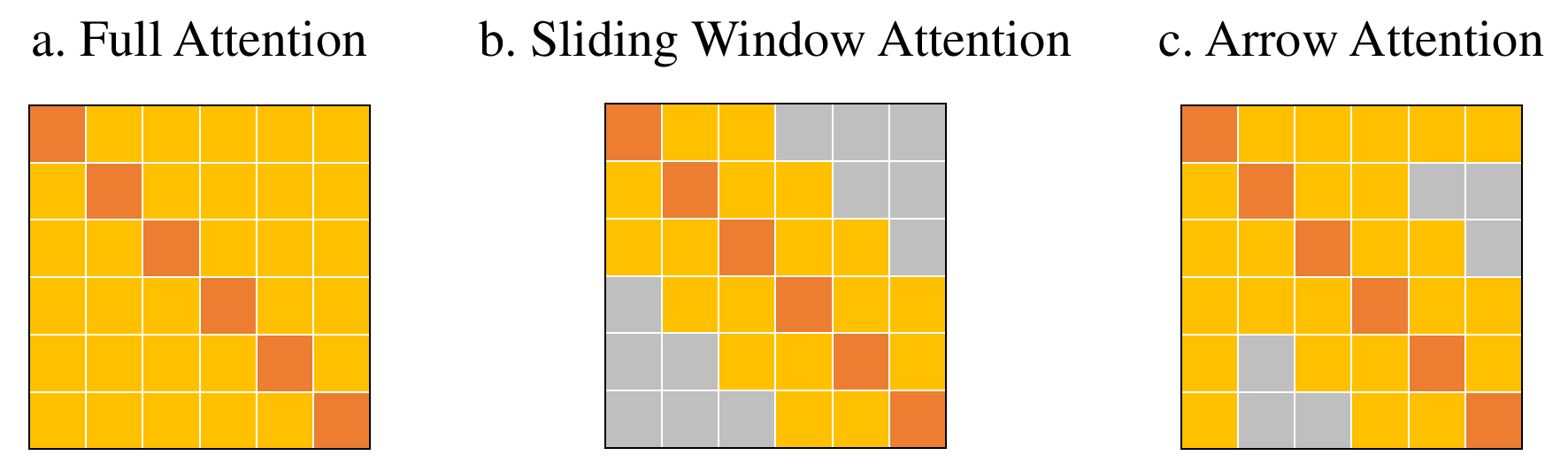} 
\caption{Illustration of arrow attention, full attention, and sliding window attention.}
\label{fig:arrow-attention}
\end{figure}

To this end, we propose an \textbf{Arrow Attention} mechanism,
where the \texttt{[CLS]} token uses global attention while other tokens use local sliding window attention, as illustrated in Figure \ref{fig:arrow-attention}.
Arrow Attention strikes a balance between global and local attention.
Compared to the computational complexity of $\mathcal{O} (L^2)$ for the full attention, 
arrow attention only requires $\mathcal{O} (wL)$.
Furthermore, the global information captured by the \texttt{[CLS]} token supplements the knowledge of SWA, 
enhancing the representation of each token and mitigating the information loss caused by the fixed receptive field.
Thus, the \texttt{[CLS]} token acts as an attention sink~\cite{xiao2023efficient} 
that balances the weights of global and local contexts.

However, varying text lengths can cause entropy instability for the {[CLS]} token, where the scale of attention scores can change significantly.
In this regard, we employ a LogN-Scaling technique on the \texttt{[CLS]} token to stabilize the entropy of attention scores.
Specifically, LogN-Scaling is defined as follows:

\begin{align}
{H}^{t}_{\texttt{[CLS]}}=\text{Attn}_{\text{s}}\left ( H^{t-1}_{\texttt{[CLS]}}W^{Q}, H^{t-1}W^{K}, H^{t-1}W^{V} \right )\\
\text{Attn}_{\text{s}}\left ( Q, K, V \right )=\text{softmax}\left ( \frac{{\log_{512}}{L}}{\sqrt{d}}QK^{\top} \right )V\label{eq3}
\end{align}

\noindent where $\text{Attn}_{s}$ is the scaled attention,
$H^{t}$ is the hidden representation of the $t$-th Transformer layer, and $W^{Q}, W^{K}, W^{V}\in \mathbb{R}^{d\times d}$ are projection matrices.

Note that LogN-Scaling is commonly used for length extrapolation in LLMs and imposed on all input tokens.
Here we utilize LogN-Scaling solely on the \texttt{[CLS]} token to improve the stability and robustness of our model.

\subsection{Biaffine Model}
Subsequently, the hidden representation $H$ is fed into a Biaffine
model to extract features for each candidate span.

\begin{align}
H^{s}, H^{e}=\text{MLP}_{\text{start}}\left ( H \right ), \text{MLP}_{\text{end}}\left ( H \right )\\
S_{i, j}=( H^{s}_i )^{\top}W_{1}H^{e}_j +W_{2} ( H^{s}_i\oplus H^{e}_j )+b\label{eq2}
\end{align}

\noindent where $\text{MLP}_{\text{start}}$ and $\text{MLP}_{\text{end}}$ are multi-layer perceptrons, 
$H^{s}/H^{e}\in \mathbb{R}^{L\times d}$ are hidden start/end embeddings, 
$W_{1}\in \mathbb{R}^{d\times c\times d}$, $W_{2}\in \mathbb{R}^{c\times 2d}$, $b\in \mathbb{R}^{c}$, 
and $c$ is the output dimension of the Biaffine model. The symbol $\oplus$ represents the concatenation operation. $S\in \mathbb{R}^{L\times L\times c}$, called \tokenspan tensor, 
denotes the hidden representation of each candidate span. 
For example, $S_{i, j}$ represents the features of $\left [ x_{i}, ..., x_{j} \right ]$.

\subsection{Token-Pair Span Interaction Module}

\begin{figure}[t]
\centering
\includegraphics[width=0.9\columnwidth]{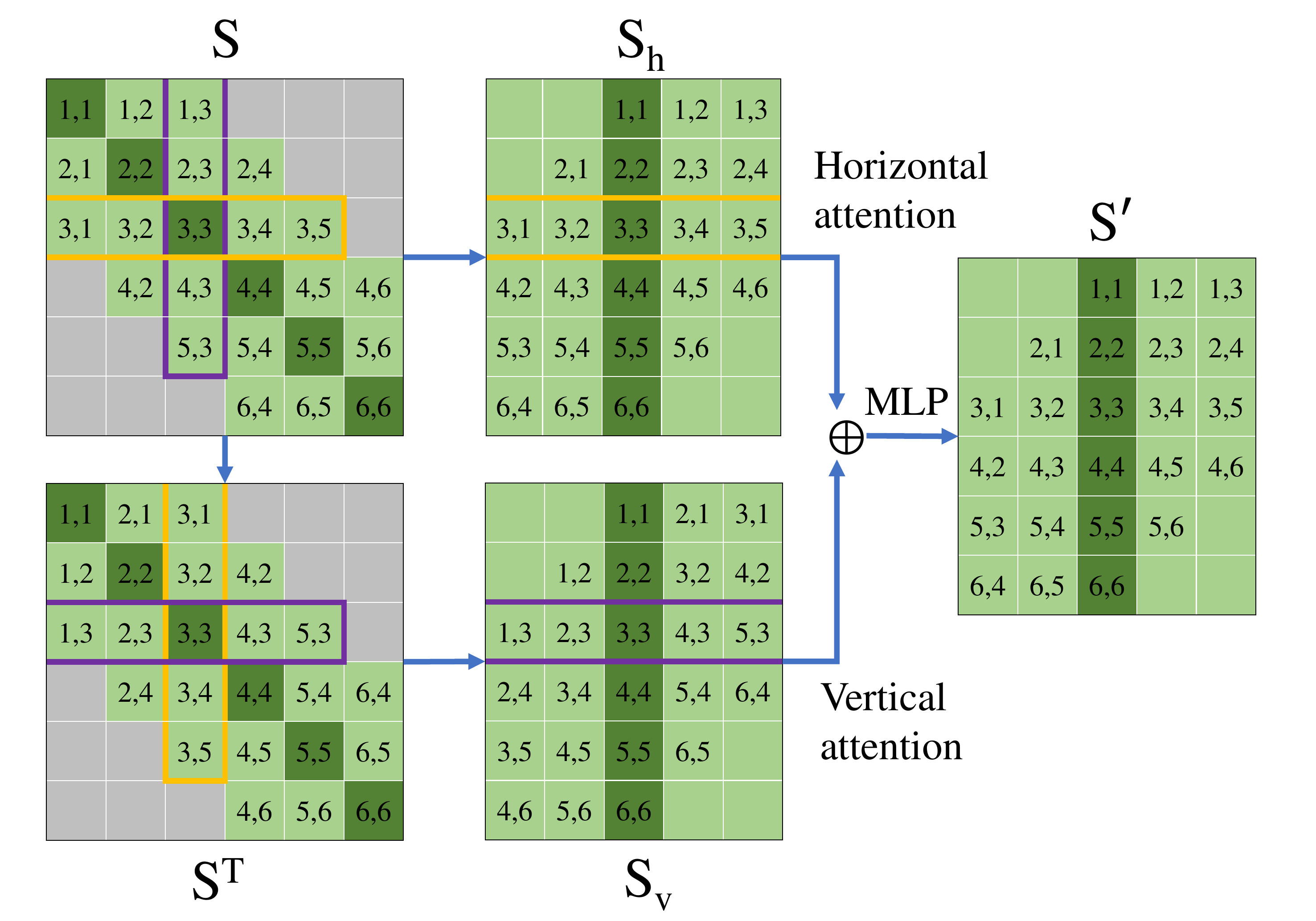} 
\caption{Diagram of the transformation for the \tokenspan tensors in \biswa mechanism.
}
\label{fig:token-pair-matrix}
\end{figure}

Note that the \tokenspan tensor  
$S$ considers each possible candidate span.
However, for long input texts,
it is unnecessary to consider every candidate span,
especially for extremely long spans.
Additionally, the GPU memory occupied by tensor $S$ increases quadratically with the input length $L$.
In light of this, we propose preserving only the hidden features of spans whose lengths do not exceed
$w'$ as shown in Figure \ref{fig:token-pair-matrix}.
Thus, $S$ is compressed to $S_{h} \in \mathbb{R}^{L\times w'\times c}$.

Previous studies~\cite{yan-etal-2023-embarrassingly,yan-etal-2023-utc} show that modeling the interactions between token pairs, such as plus-shaped and local interaction, should be helpful.
Plus-shaped attention applies the self-attention mechanism horizontally and vertically.
However, plus-shaped attention cannot be performed directly on the compressed hidden feature tensor $S_{h}$ since either the original horizontal or vertical dimension is disrupted.
Therefore, we propose a novel bidirectional sliding-window plus attention (\textbf{BiSPA}) mechanism to perform plus-shaped attention on the compressed $S_{h}$.

Specifically, we first compute the horizontal self-attention on $S_{h}$,
as shown in the top middle of Figure \ref{fig:token-pair-matrix}.
Next, we propose a transformation method,
that transforms the top left matrix $S$ to the bottom middle matrix $S_{v}$,
and then compute the vertical self-attention based on $S_{v}$.
Finally, we concatenate the horizontal and vertical attention matrices and feed them into an MLP to aggregate plus-shaped perceptual information.
Notably, the computational complexity of the BiSPA mechanism is reduced from $\mathcal{O}(L^3)$ to $\mathcal{O}(L \times (w^{'})^2)$,
optimizing the 
training efficiency significantly.

\begin{align}
Z^{h/v}_{i,:} =\text{Attn}\left ( S^{h/v}_{i,:}W_{h/v}^{Q}, S^{h/v}_{i,:}W_{h/v}^{K}, S^{h/v}_{i,:}W_{h/v}^{V} \right )\\
\text{Attn}\left ( Q, K, V \right )=\text{softmax}\left ( \frac{QK^{T}}{\sqrt{c}} \right )V\label{eq3}
\end{align}

\begin{equation}
{S}'=\text{MLP}\left ( Z^{h}\oplus Z^{v} \right )\label{eq4}
\end{equation}

\noindent where $W_{h}^{Q}, W_{h}^{K}, W_{h}^{V}$, $W_{v}^{Q}, W_{v}^{K}, W_{v}^{V}\in \mathbb{R}^{c\times c}$, $Z^{h}/Z^{v}$ is intermediate representation after horizontal/vertical self-attention, 
and ${S}'\in \mathbb{R}^{L\times w'\times c}$ is the \tokenspan feature after \biswa mechanism.

The BiSPA mechanism endows the model with the capacity to perceive horizontal and vertical directions.
We further use two types of position embeddings to enhance the sense of  distances between token pairs and the area the token pair locates~\cite{yan-etal-2023-utc}. (1) Rotary Position Embedding (RoPE)~\cite{su2024roformer} 
encodes the relative distance between token pairs,
which is used for both horizontal and vertical self-attention.
(2) Matrix Position Embedding indicates whether each entry in $S'$ is the original upper or lower triangles,
which adds to $S_{h}$ and $S_{v}$.

After the BiSPA mechanism, we employ  CNN 
with kernel size $3 \times 3$ on $S^{'}$ to model the local interactions between \tokenspans.

\begin{equation}
{S}''= \text{Recover} ( \text{Conv}\left ( \sigma \left ( \text{Conv}\left ( {S}' \right ) \right ) \right ) ) \label{eq5}
\end{equation}

\noindent where ${S}''\in \mathbb{R}^{L\times L \times c}$ is recovered to the square size,
and $\sigma $ is the activation function. 

We name the module encompassing the BiSPA mechanism and the convolutional module as the BiSPA Transformer block.
The BiSPA Transformer blocks will be repeatedly used to ensure full interaction between token pairs.

\subsection{Training and Prediction}

We utilize MLP layers to transform the output of the final BiSPA Transformer block into output scores. We use binary cross-entropy as the loss function.

\begin{equation}
\widehat{Y}=\text{MLP}\left ( {S}''+ S \right )\label{eq6}
\end{equation}

\begin{align}
\mathcal{L}= &-\sum_{i,j=1}^{L}\sum_{r=1}^{R}\left ( Y_{i,j}^{r}log\left ( \widehat{Y}_{i,j}^{r} \right ) \right.\\
&\left. +\left ( 1-Y_{i,j}^{r} \right )log\left ( 1-\widehat{Y}_{i,j}^{r} \right ) \right )
\label{eq7}
\end{align}

\noindent where $\widehat{Y}\in \mathbb{R}^{L\times L\times R}$ represents the scores of candidate entities, and $R$ is the number of entity types.

To improve the robustness and generalization of our model,
we employ the whole word masking strategy~\cite{cui2021pre} during training and utilize LoRA~\cite{hu2021lora} technique to train the PLM parameters.

During prediction, our model uses the average of the upper triangular and lower triangular values as the final prediction score, as follows:

\begin{equation}
P_{i,j}^{r}= \frac{\left ( \widehat{Y}_{i,j}^{r}+\widehat{Y}_{j,i}^{r} \right )}{2}, i\leq j\label{eq8}
\end{equation}

All text spans that satisfy $P_{i,j}^{r}>0$ 
are outputted. 
If the boundaries of multiple candidate spans conflict, the span with the highest prediction score is selected.

%% file: Experiment.tex
\section{Experiment}

\subsection{Datasets}

We conduct experiments on three NER datasets: \profiling~\cite{zhang2024oag}, \scirex~\cite{jain-etal-2020-scirex}, and \oldprofiling~\cite{tang2007social,tang2008arnetminer}. The statistics of all datasets are detailed in Table \ref{statistics}. 
As shown in Table \ref{statistics}, the input lengths and entity lengths of the three datasets are longer than those of traditional named entity recognition datasets, presenting greater challenges. 

\begin{table}[t]
        \newcolumntype{C}{>{\centering\arraybackslash}p{3.2em}}
        \small
        \centering
        \renewcommand\arraystretch{1.0}
        \begin{tabular}{c|@{~ }*{1}{ccc}}
            \toprule[1.2pt]
        {}   & {\profiling}  & {\scirex}  & {\oldprofiling} \\
        \midrule
        Input avg. len. & 433.42 & 5678.29 & 785.09 \\
        Input max. len. & 692 & 13731 & 17382 \\
		Input num. & 2099 & 438 & 1446 \\
        Entity num. & 20994 & 156931 & 17416 \\
        Entity type & 12 & 4 & 13 \\
        Entity avg. len. & 12.45 & 2.28 & 8.88  \\
        Entity max. len. & 480 & 18 & 307  \\
        \bottomrule[1.2pt]
    \end{tabular}
    \caption{
            \label{statistics} Statistics of the datasets (in words). 
        }
\end{table}

\subsection{Baselines}

We compare our model with several recent NER methods:

Span-based Methods: 
\textbf{CNN-NER}~\cite{yan-etal-2023-embarrassingly}: is a span-based method that utilizes Convolutional
Neural Networks (CNN) to model local spatial correlations between spans.
\textbf{UTC-IE}~\cite{yan-etal-2023-utc}: models axis-aware interaction with plus-shaped self-attention and local interaction with CNN on top of the \tokenspan tensor.

Others Methods: 
\textbf{DiffusionNER}~\cite{shen2023diffusionner}: formulates the NER task as a boundary-denoising diffusion process and thus generates named entities from noisy spans.

Generation-based Methods: 
\textbf{UIE}~\cite{lu2022unified}: uniformly encodes different extraction structures via a structured extraction language, adaptively generates target extractions, and captures the common IE abilities via a large-scale pre-trained text-to-structure model.
\textbf{InstructUIE}~\cite{wang2023instructuie}: leverages natural language instructions and instruction tuning to guide large language models for IE tasks.
\textbf{GOLLIE}~\cite{sainz2023gollie}: is based on Code-Llama~\cite{roziere2023code} and fine-tunes the foundation model to adhere to specific annotation guidelines.
\textbf{ADELIE}~\cite{qi2024adelie}: 
builds a high-quality instruction tuning dataset and utilizes supervised fine-tuning (SFT) followed by direct preference optimization (DPO).
\textbf{ToNER}~\cite{jiang2024toner}: firstly employs an entity type matching model to discover the entity types that are most likely to appear in the sentence, 
and then adds multiple binary classification tasks to fine-tune the encoder in the generative model.
\textbf{GPT-4o}~\cite{achiam2023gpt}: employs the gpt-4o-2024-08-06 API, utilizing a 5-shot in-context learning approach to enhance performance.
\textbf{Claude-3.5}~\cite{TheC3}: uses the claude-3-5-sonnet-20241022 API, also adopting 5-shot in-context learning.

\subsection{Experimental Setup}

All experiments are conducted on an 8-card 80G Nvidia A100 server. 
The entire text is used for the \profiling dataset, while the other two datasets are truncated to $5120$ using a sliding window approach, as a trade-off due to limited GPU memory.
For prediction, 
the prediction of the text segment is mapped to the starting/ending position of the original text.
Hyper-parameters are selected based on the F1 score on the validation set.
For each experiment, we run $3$ times with different random seeds and report the average results.
We choose DeBERTa-V3-large~\cite{he2023debertav3improvingdebertausing} as the PLM for span-based methods and DiffusionNER. 
We use AdamW~\cite{loshchilov2017fixing} optimizer with a weight decay of $1e^{-2}$. 
The unilateral window sizes of the arrow attention and BiSPA mechanism are both set to $128$. 
We only use low-rank adaptation on the $Q$ and $V$ matrix of the self-attention mechanism with a rank of $8$. 

\subsection{Evaluation Metrics}

We report the micro-F1 score for all attributes. 
An entity is considered correct only if both the entity type and the entity span are predicted correctly.
Precision (P) is the portion of correctly predicted spans over predicted spans, while Recall (R) is the portion of correctly predicted spans over ground-truth entity spans.

\subsection{Main Results}

\begin{table*}[t]
        \newcolumntype{C}{>{\centering\arraybackslash}p{2.2em}}
        \centering
        \renewcommand\arraystretch{1.0}
        \begin{tabular}{c|c|@{~ }*{1}{CCC|}*{1}{CCC|}*{1}{CCC}}
            \toprule[1.2pt]
        \multirow{2}{*}{Method type}    
            &\multirow{2}{*}{Method} 
            &\multicolumn{3}{c|}{\profiling}
            &\multicolumn{3}{c|}{\scirex}
            &\multicolumn{3}{c}{\oldprofiling}
        \\
        \cmidrule{3-5} \cmidrule{6-8}  \cmidrule{9-11} 
        & & {P} & {R} & {F1}  & {P} & {R} & {F1}  & {P} & {R} & {F1} \\
        \midrule
        \multirow{7}{*}{\shortstack{Generation-based\\Methods}}
        & UIE & 43.32 & 36.80 & 39.80 & 65.88 & 56.44 & 60.80 & 65.92 & 57.51 & 61.43 \\
		& ToNER & 40.08 & 29.48 & 33.97 & 57.43 & 31.56 & 40.73 & 48.80 & 41.21 & 44.68  \\
        & InstructUIE & 34.63 & 36.50 & 35.54 & 56.31 & 54.60 & 55.44 & 59.09 & 63.19 & 61.07  \\
        & GOLLIE & 43.74 & 40.88 & 42.26 & 71.56 & 71.50 & 71.53 & 64.51 & 9.46 & 16.50 \\
        & ADELIE & 45.60 & 39.05 & 42.07 & 70.10 & 71.84 & 70.96 & 65.75 & 49.53 & 56.50  \\
        & Claude-3.5 & 17.15 & 30.16 & 21.87 & 57.78 & 7.97 & 14.01 & 34.40 & 43.07 & 38.25  \\
        & GPT-4o & 18.24 & 27.31 & 21.87 & 40.44 & 7.69 & 12.92 & 36.96 & 43.73 & 40.06  \\
        \midrule
        \multirow{1}{*}{Others Methods}
        & DiffusionNER & \underline{55.33} & 29.87 & 38.80 & \textbf{77.11} & 62.36 & 68.96 & \textbf{70.51} & 44.16 & 54.31   \\
        \midrule
        \multirow{3}{*}{\shortstack{Span-based\\Methods}}
        & CNN-NER & 50.92 & 44.72 & 47.59 & 72.13 & 74.56 & 73.32 & 69.19 & 62.79 & 65.56 \\
        & UTC-IE & 53.17 & \underline{46.01} & \underline{49.10} & 71.90 & \underline{75.09} & \underline{73.42} & \underline{69.79} & \underline{65.28} & \textbf{67.43} \\
        & \model (Ours) & \textbf{57.41} & \textbf{46.80} & \textbf{51.56} & \underline{72.89} & \textbf{76.17} & \textbf{74.49} & 67.52 & \textbf{67.17} & \underline{67.34} \\
        \bottomrule[1.2pt]
    \end{tabular}
    \caption{
            \label{performance}  Main results on three long NER datasets (\%). 
            The best results are \textbf{boldfaced} and the second best results are \underline{underlined}.
        }
\end{table*}

Table \ref{performance} provides a holistic comparison of different NER methods on three datasets.
Generally speaking, span-based methods (CNN-NER, UTC-IE, and our model \model) outperform other types of NER methods.

Generation-based methods utilize generation loss to fine-tune the foundation model to adapt to the long NER task, achieving unfavorable performance.
UIE outperforms InstructUIE,
possibly because UIE defines a structured extraction language that suits the long NER problem better than naively performing instruction tuning.
GOLLIE and ADELIE achieve similar performance, except for \oldprofiling dataset,
which is due to the fact that this dataset is sliced and diced with a high number of empty data and thus makes GOLLIE overfit these empty examples.
ToNER obtains unsatisfactory performance,
possibly since the two-stage framework leads to error propagation
and the usage of small language models for generation limits its potential.
GPT-4o and Claude-3.5-sonnet are less effective, suggesting that the proprietary model prompting does not perform the long text NER task well.

The span-based NER methods (CNN-NER, UTC-IE, and \model) outperform other types of NER methods, including Diffusion-NER.
DiffusionNER is a diffusion-based method that recovers the boundaries of the entities from a fixed amount of Gaussian noise
and it is hard to recover longish entities from long texts.
CNN-NER models fine-grained span interactions via CNN,
achieving decent extraction performance.
UTC-IE further improves CNN-NER by introducing plus-shaped attention on the \tokenspan tensor,
achieving consistent outperformance over CNN-NER.

Our model \model exhibits noticeable improvements or is on par with the best baseline,
suggesting that with the design of arrow attention coupled with LogN-Scaling on the \texttt{[CLS]} in the PLM encoder, as well as the BiSPA mechanism on the \tokenspan tensor,
our model is capable of saving computation and memory resources without degrading the extraction accuracy.
In addition, longer text with more focused attention can effectively help the model understand the semantic information of the text in more detail and extract the corresponding entities.

\subsection{Ablation Study}

\begin{table}[t]
\small
        \newcolumntype{C}{>{\centering\arraybackslash}p{2.2em}}
        \centering
        \renewcommand\arraystretch{1.0}
        \begin{tabular}{c|@{~ }*{1}{CC|}*{1}{CC|}*{1}{CC}}
            \toprule[1.2pt]

        {Method} & \multicolumn{2}{c|}{\profiling} & \multicolumn{2}{c|}{\scirex} & \multicolumn{2}{c}{\oldprofiling} \\
        \cmidrule{2-3} \cmidrule{4-5}  \cmidrule{6-7} 
        & {F1} & {Mem} & {F1} & {Mem}  & {F1} & {Mem} \\
        \midrule
        \model & \textbf{51.56} & 16.95 & \textbf{74.49} & 69.23 & \textbf{67.34} & 63.36 \\
        \text{w/o Arrow} & 51.34 & 16.94 & - & OOM & - & OOM \\
        \text{w/o LogN} & 50.78 &17.18 & 74.27 & 68.71 & 67.11 & 63.47 \\
        \text{w SWA} & 49.95 & 16.46 & 73.50 & 68.33 & 65.60 & 63.16 \\
        \text{w/o LoRA} & 49.93 & 17.98 & 73.98 & 70.79 & 66.53 & 66.58 \\
        \text{w/o \biswa} & 50.48 & 41.39 & - & OOM & - & OOM  \\
        \text{w/o WWM} & 50.45 & 17.14 & 74.24 & 69.00 & 64.97 & 63.37 \\
        \bottomrule[1.2pt]
    \end{tabular}
    \caption{
            \label{ablation}  Ablation studies on three long NER datasets. Mem means memory usage (GB), SWA denotes sliding window attention and WWM is whole word masking.
        }
\end{table}

Table \ref{ablation} presents a justification for the effectiveness of
each component in our model.
Removing either the arrow attention or \biswa mechanism results in a decrease in model performance on \profiling and Out-of-Memory (OOM) errors on \scirex and \oldprofiling. 
It indicates that both modules effectively reduce explicit memory usage, enabling the model to handle longer texts and thereby improving overall performance.
Specifically, \biswa significantly reduces compute and memory footprint by reducing negative samples.
In contrast, the arrow attention has a limited ability to reduce memory usage for short text on the \profiling dataset.
Substituting the arrow attention with sliding window attention (SWA) leads to a significant performance drop, highlighting the necessity of 
imposing attention scores on the \texttt{[CLS]} token  to absorb global contextual information.
Adding LogN-Scaling consistently improves the performance, 
thereby enhancing model stability and robustness. 
Although removing LoRA does not cause OOM errors, the F1 score decreases across all datasets to some extent, demonstrating that LoRA can effectively reduce training parameters and prevent overfitting. 
Whole Word Masking (WWM) increases the diversity of input texts,
thus improving the generalization capacity of the model.

\subsection{Detailed Analysis for Entity Types}

In this subsection, we focus on comparing the performance of our method with span-based methods (CNN-NER and UTC-IE) and LLM-based methods (InstructUIE, GOLLIE, and ADELIE) across entities of varying lengths and types.

\begin{figure}[t]
\centering
\includegraphics[width=0.9\columnwidth]{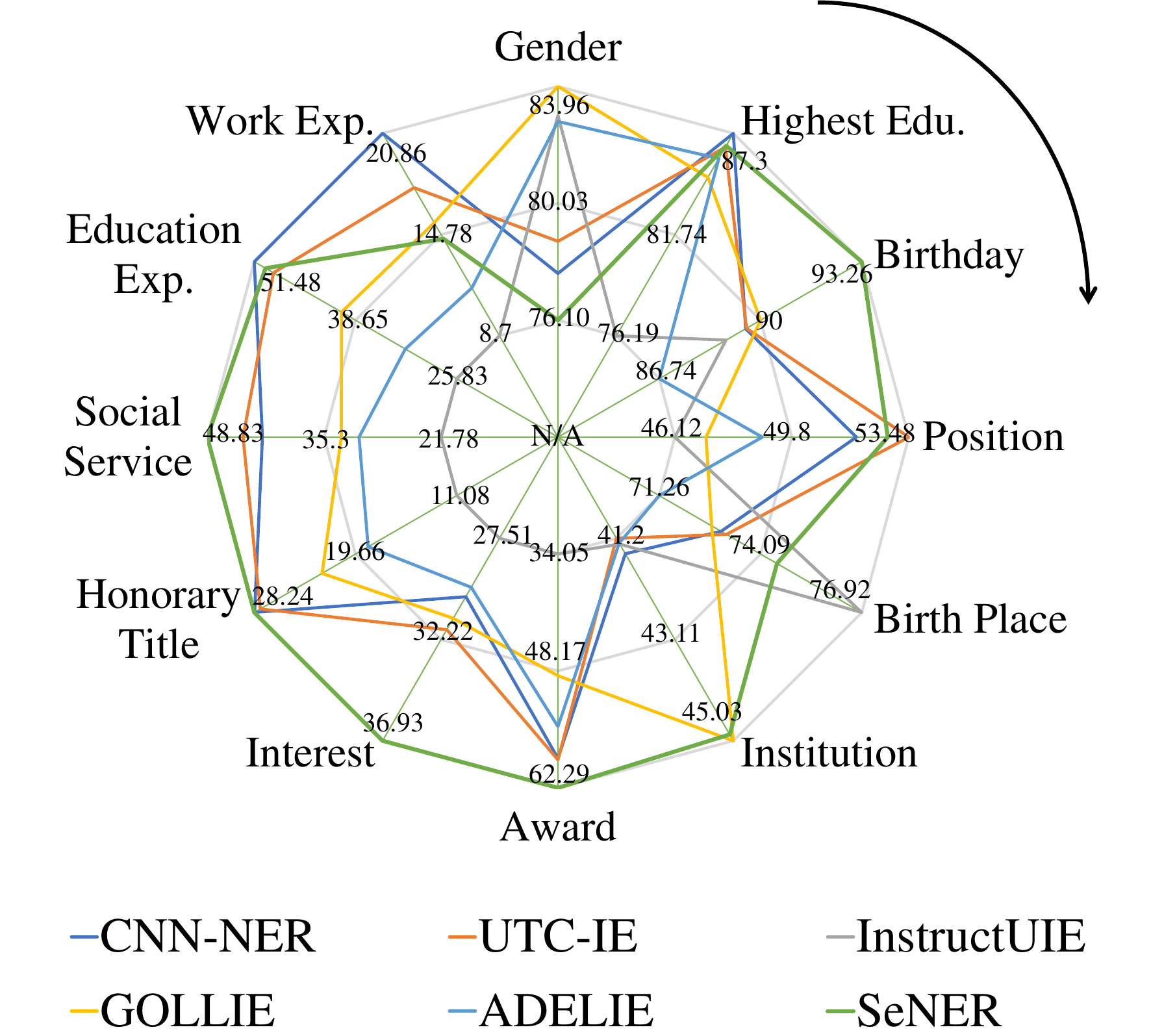} 
\caption{Performance of different entity types on the \profiling dataset (\%).
The average length of entities increases clockwise from Gender to Work Exp. (Highest Edu.: Highest Education, Education Exp.: Education Experience, Work Exp.: Work Experience.)
}
\label{fig:radar-profiling}
\end{figure}

The results on the \profiling dataset are depicted in Figure \ref{fig:radar-profiling},
with the average length of the entity types increasing clockwise from ``Gender'' to ``Work Experience''. 
Generative methods, leveraging the powerful capabilities of LLMs, achieve superior performance in extracting ``Gender'' and ``Birth Place'' types. 
However, for other types of entities, span-based methods demonstrate consistent superiority. 
Our model \model
outperforms CNN-NER and UTC-IE in most types of entities, with particularly notable improvements for longish entities. 
Specifically, 
for ``Social Service'',
our method achieves an improvement of $6.38\%$ over CNN-NER and $4.14\%$ over UTC-IE, respectively. 
The performance of \model for entity types ``Education Experience'' and ``Work Experience'' falls behind the leading ones a little,
indicating that the approximation strategy in our model inevitably loses some information, especially on very long entities. 

\subsection{Analysis for Maximum Input Length}

\begin{figure}[t]
\centering
\includegraphics[width=0.95\columnwidth]{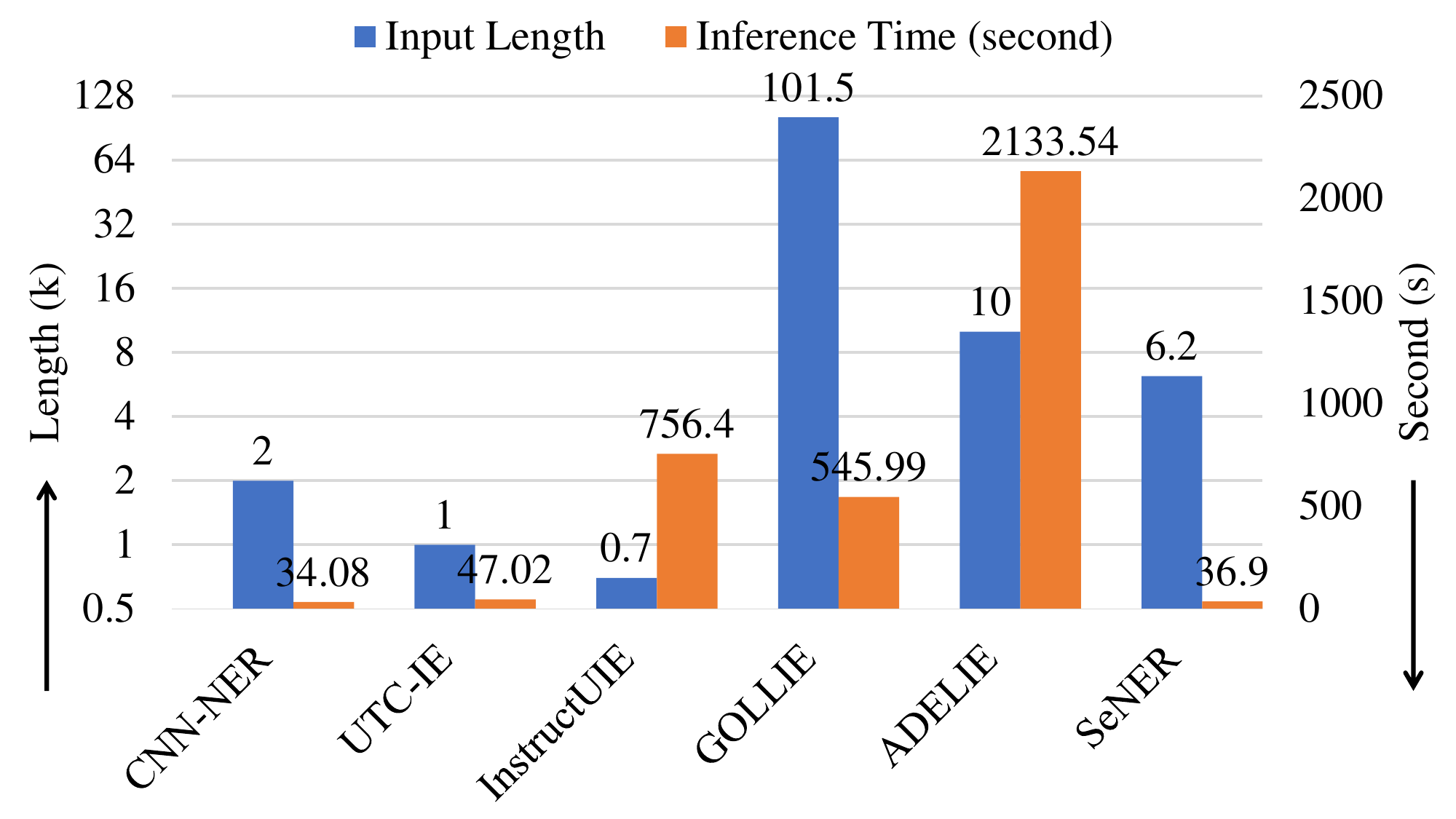} 
\caption{
Blue bar: Maximum input length comparison of different methods (k).
Orange bar: Inference time comparison of different methods (second).
Both are conducted on the longest \scirex dataset.
}
\label{fig:memory-analysis}
\end{figure}

We examine the maximum input length supported by training each NER method
on a single Nvidia A100 with a batch size of $1$,
as shown in Figure \ref{fig:memory-analysis}. 
Generation-based methods 
often employ various lightweight strategies, such as quantization,  FlashAttention~\cite{dao2022flashattention}, and Zero Redundancy Optimizer (ZeRO)~\cite{rajbhandari2020zero}
enabling models like GOLLIE and ADELIE to handle long texts. 
In contrast, 
span-based methods need to model \tokenspan tensor, 
resulting in supporting shorter input length.
Our method, \model, demonstrates substantial improvements over CNN-NER and UTC-IE, supporting input lengths that are $3$ times and $6$ times longer, respectively.

\subsection{Efficiency Performance for Inference Time}
Figure \ref{fig:memory-analysis} also displays the inference time of span-based methods and LLM-based methods.
It can be observed that LLM-based methods lead to $10$ times longer inference time than span-based methods.
Our method \model achieves a similar inference time compared with CNN-NER, achieving significantly better extraction accuracy simultaneously.
\model can save $20\%$ inference time compared with UTC-IE and encode longer input texts, still maintaining state-of-the-art extraction accuracy.

%% file: Conclusion.tex
\section{Conclusion} 
In this paper, we tackle the problem of extracting entities from long texts, a less explored area in Named Entity Recognition (NER).  
Current span-based and generation-based NER methods face issues such as computational inefficiency and memory overhead in span enumeration, along with inaccuracy and time costs in text generation. 
To address these challenges, we introduce \model, a lightweight span-based approach that featuring a bidirectional arrow attention mechanism and LogN-Scaling for effective long-text embedding. 
Additionally, we propose a bidirectional sliding-window plus-shaped attention (\biswa) mechanism that significantly reduces redundant candidate \tokenspans and models their interactions. 
Extensive experiments show that SeNER achieves state-of-the-art accuracy in extracting entities from long texts across three NER datasets, while maintaining GPU-memory efficiency.
 Our innovations in arrow attention and the \biswa mechanism have the potential to advance future research in information extraction tasks.

%% file: appendix.tex
\section{Hyper-parameter Search}

We optimize the model's hyper-parameters based on its performance metrics on the validation set, and then evaluate its final performance on the test set. The parameter search range is as follows: the learning rate search range includes $2e^{-4}$, $3e^{-4}$, $4e^{-4}$; the unilateral window size for the arrow attention and \biswa mechanisms includes 32, 64, 128, 256, 512; and the masking strategy options are token masking, whole word masking, and span masking.

\section{Prompt Construction}

On the three datasets, we prompt GPT-4o and Claude-3.5 to extract entities by providing five similar demonstrations. The format of the prompt is as follows:

Recognize entities from the following sentence and classify the entity type into the options. Options: [type 1, type 2, ..., type r]. Please give the answer in json format.
{example 1}, {example 2}, ...,  {example 5}. Text: .... Output: 

\section{Detailed Experimental Setup}

For the baseline methods, to balance GPU memory usage and training time overhead, the maximum length of the input text is set to 512. A sliding window of 512 is used to segment the text, and the results of these segments are integrated during prediction. The remaining hyper-parameters are determined through searching on the validation set, and the optimal ones are selected. Since some methods require annotation guidelines on entity types (GOLLIE and ADELIE) as complementary knowledge, we use GPT-4o to generate five detailed descriptions for each entity type.

\section{Entity Types}

The three datasets contain the following entity types, arranged from smallest to largest based on average length.The numbers in parentheses represent the average length of entities of the corresponding type, counted in words.

The Scholarst-XL dataset contains 12 entity types: Gender (1), Highest Education (1.07), Birthday (2.43), Position (2.49), Birth Place (2.67), Institution (5.21), Award (7.58), Interest (8.53), Honorary Title (8.83), Social Service (11.12), Educational Experience (40.34), and Work Experience (56.33). 

The SciREX dataset contains 4 entity types: Metric (1.95), Material (2.02), Method (2.35), and Task (2.44).

The Profiling-07 dataset contains 13 entity types: Date (1.38), Major (2.18), Position (2.19), Degree (3.2), Univ (3.41), Interest (4.85), Phone (6.28), Fax (6.34), Affiliation (7.07), Email (7.41), Address (9.61), Contact\_info (44.78), and Education\_info (78.59). Here, Date and Univ denote the date and university of graduation, respectively. Contact\_info and Education\_info represent contact information and educational experience, respectively.

\section{Window Size Sensitivity}

\begin{figure}[t]
\centering
\includegraphics[width=0.85\columnwidth]{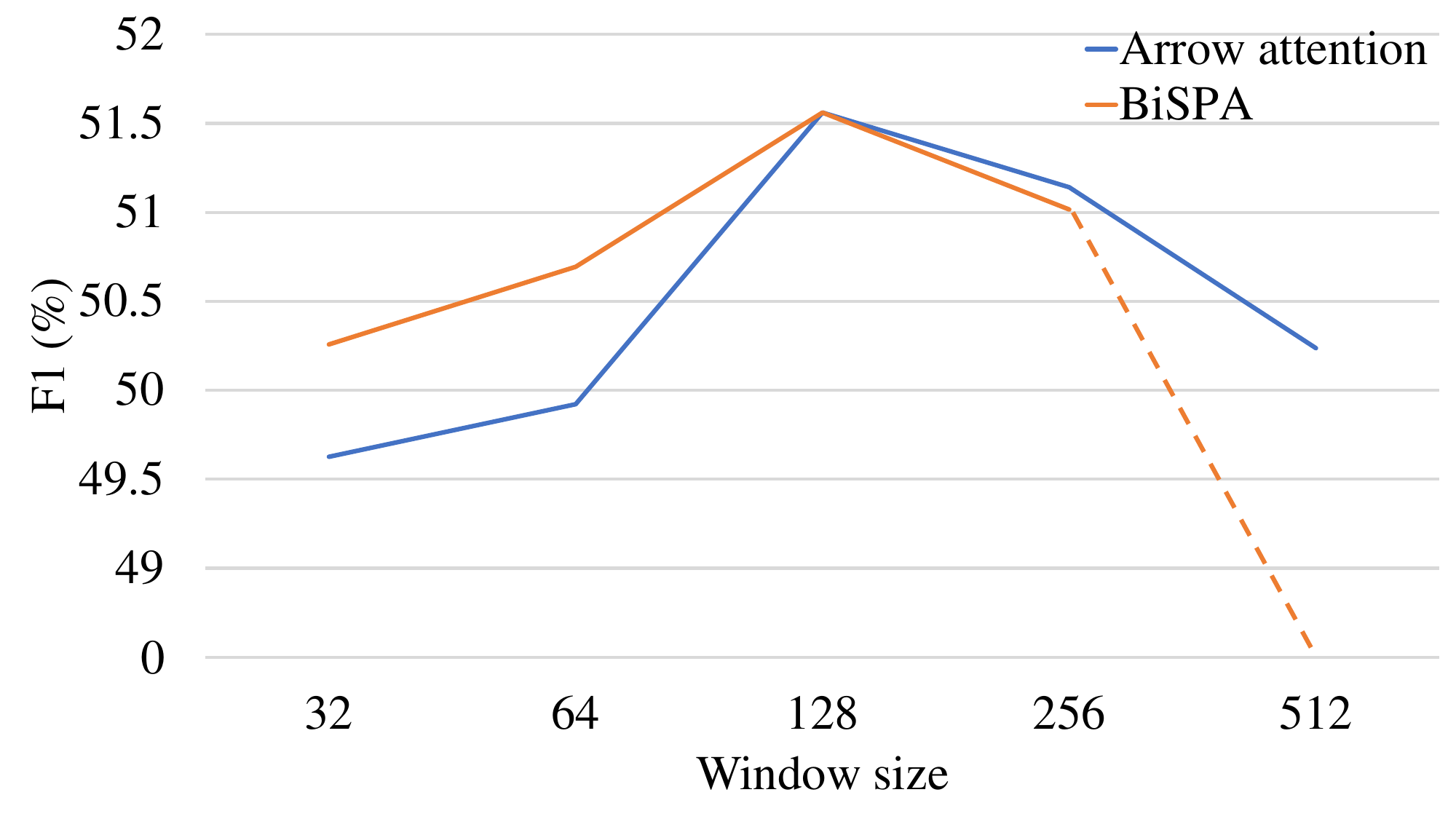} 
\caption{
Impact of window sizes of arrow attention and \biswa mechanism on the \profiling dataset (\%).
}
\label{fig:window-size-exploration}
\end{figure}

We investigate how the unilateral window size of the arrow attention and \biswa mechanism impacts the performance, as shown in Figure \ref{fig:window-size-exploration}. 
For the arrow attention, a small window restricts the information aggregation capability of local attention, leading to the loss of critical information. 
Conversely, an excessively large window increases the difficulty in information focusing, 
resulting in degraded performance. 

For the \biswa mechanism, a small window reduces the number of candidate entities in the \tokenspan tensor, making it difficult to extract long entities effectively. 
On the other hand, a large window retains a large number of redundant candidate entities, introduces more false positives, and consumes significant computational resources. 
Additionally, when the window size is set to $512$, an Out-of-Memory (OOM) error occurs, further demonstrating the effectiveness of the \biswa mechanism.